\documentclass[10pt,twocolumn,letterpaper]{article}

\usepackage[pagenumbers]{wacv} 

\usepackage{graphicx}
\usepackage{amsmath}
\usepackage{amssymb}
\usepackage{booktabs}

\usepackage{bm}
\usepackage{enumitem}
\usepackage[dvipsnames]{xcolor}
\usepackage{xspace}

\usepackage[pagebackref,breaklinks,colorlinks]{hyperref}
\usepackage[labelfont=bf]{caption}

\usepackage[capitalize]{cleveref}
\crefname{section}{Sec.}{Secs.}
\Crefname{section}{Section}{Sections}
\Crefname{table}{Table}{Tables}
\crefname{table}{Tab.}{Tabs.}

\definecolor{MyDarkBlue}{rgb}{0,0.08,1}
\definecolor{MyDarkGreen}{rgb}{0.02,0.6,0.02}
\definecolor{MyDarkRed}{rgb}{0.8,0.02,0.02}
\definecolor{MyDarkOrange}{rgb}{0.40,0.2,0.02}
\definecolor{MyPurple}{RGB}{111,0,255}
\definecolor{MyRed}{rgb}{1.0,0.0,0.0}
\definecolor{MyGold}{rgb}{0.75,0.6,0.12}
\definecolor{MyDarkgray}{rgb}{0.66, 0.66, 0.66}
\definecolor{MyWineRed}{rgb}{0.694,0.071, 0.149}
\definecolor{nicegreen}{rgb}{0.1, 0.6, 0.2}
\definecolor{JiayuanColor}{rgb}{0.60,0.43,0.48}

\def\Tabref#1{Table~\ref{#1}}

\newcommand{\model}{ShapeCraft\xspace}

\def\Figref#1{Figure~\ref{#1}}

\def\eqref#1{equation~\ref{#1}}

\def\1{\bm{1}}

\DeclareMathAlphabet{\mathsfit}{\encodingdefault}{\sfdefault}{m}{sl}
\SetMathAlphabet{\mathsfit}{bold}{\encodingdefault}{\sfdefault}{bx}{n}

\def\wacvPaperID{1114} 
\def\confName{WACV}
\def\confYear{2025}

\begin{document}

\title{CRAFT: Designing Creative and Functional 3D Objects}
\author{
\begin{tabular}{cccc}
Michelle Guo* & Mia Tang* & Hannah Cha & Ruohan Zhang\\
 & C. Karen Liu & Jiajun Wu
\end{tabular}
\\[2ex]
Stanford University\\[1ex]
{\tt\small \{mguo95,miatang,hcha417,zharu,karenliu,jiajunwu\}@cs.stanford.edu}
}
\twocolumn[{%
\renewcommand\twocolumn[1][]{#1}%
\maketitle
\begin{center}
    \centering
    \captionsetup{type=figure}
    \includegraphics[width=0.99\linewidth]{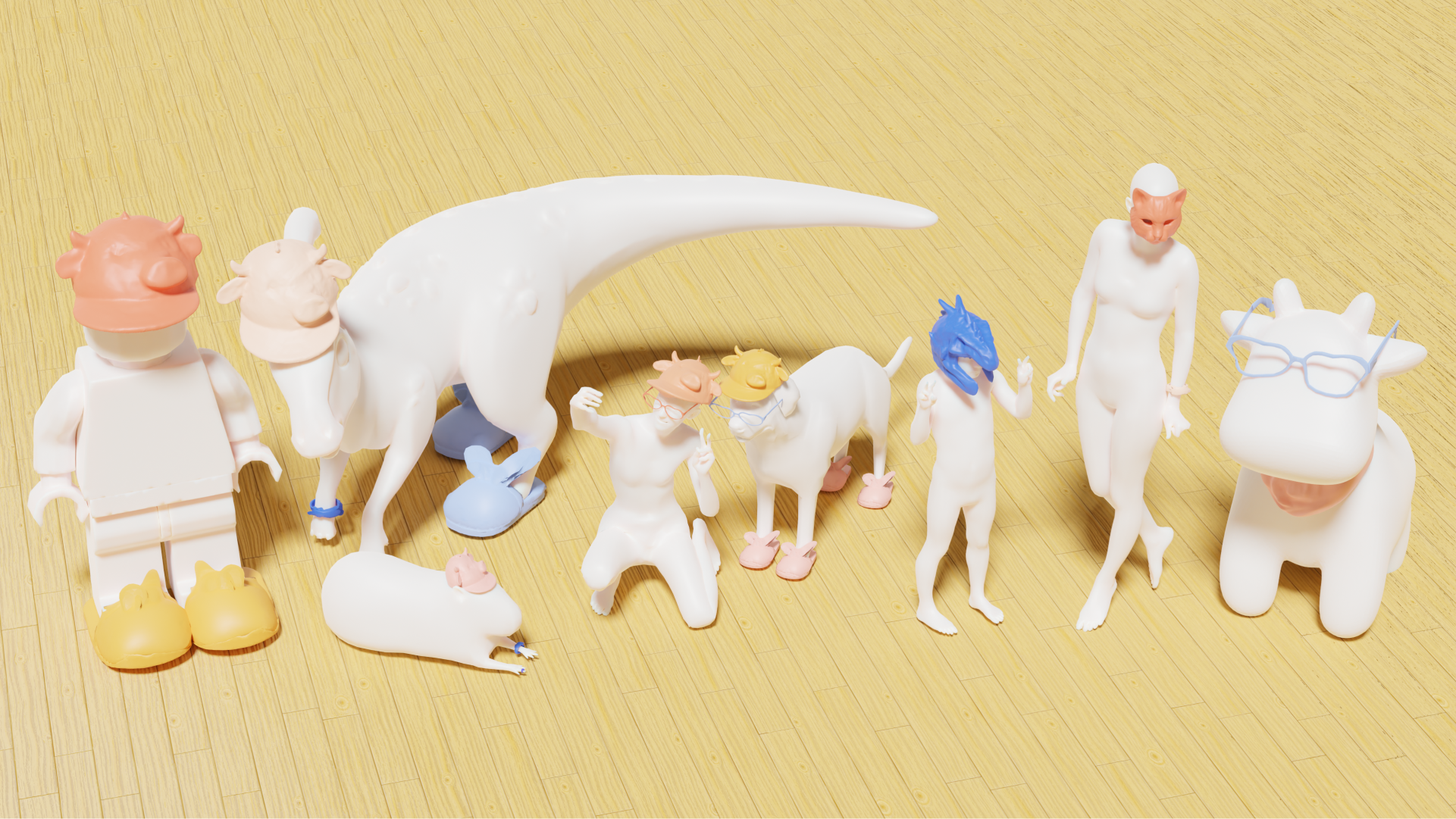}
    \captionof{figure}{\model generates 3D shapes given text as input. The objects are optimized to fit on various character body shapes.}
    \vspace{10pt}
    \label{fig:teaser}
\end{center}%
}]


\begin{abstract}
For designing a wide range of everyday objects, the design process should be aware of both the human body and the underlying semantics of the design specification. However, these two objectives present significant challenges to the current AI-based designing tools. In this work, we present a method to synthesize body-aware 3D objects from a base mesh given an input body geometry and either text or image as guidance. The generated objects can be simulated on virtual characters, or fabricated for real-world use. We propose to use a mesh deformation procedure that optimizes for both semantic alignment as well as contact and penetration losses. Using our method, users can generate both virtual or real-world objects from text, image, or sketch, without the need for manual artist intervention. We present both qualitative and quantitative results on various object categories, demonstrating the effectiveness of our approach.
\end{abstract}

\section{Introduction}
\label{sec:intro}

Have you struggled to find an everyday object that will fit your body perfectly and match the exact creative concept you have in mind?  Recent progress in generative AI models shows promising results in generating 3D objects, which have the potential to facilitate the design process (e.g., help designers rapidly iterate ideas) and enable better customization in industrial design \cite{epstein2023art,chui2023economic,makatura2023can}.
For designing a wide range of everyday objects, such as glasses, hats, rings, and shoes, the designing process should be aware of both the human \emph{body} and the object \emph{semantics}. 
For these objects that are designed to be used by humans, being body-aware is essential and the design should be primarily optimized for the interaction between the object and the body that it is designed for. In addition, we want to be able to customize the design, styles, or aesthetics of these objects, i.e., we want the design to be semantically-aware in the sense that it aligns with our design specifications, which can be either text descriptions or visual examples. Therefore, we need to provide tools to address individual differences in the different object categories' demands of body fit and the underlying semantics of the designs.

Generative AI models, such as Stable Diffusion~\cite{rombach2022high}, DALL-E~\cite{dalle3}, and DreamFusion~\cite{poole2022dreamfusion}, can generate semantics-aware 3D assets when given design specifications in the format of natural language, although text-to-image models would require another step that converts 2D designs into 3D objects using image-to-3D models~\cite{liu2023zero}. However, these approaches typically optimize objects for semantics-related objectives, such as prompt alignment. Meanwhile, designing useful objects requires an understanding of the physical interactions between bodies and objects. It is difficult to use text or image to specify design needs for different body shapes and preferred body contacts, hence the resulted designs are not sufficiently body-aware. Additionally, the generated designs are derivatives of datasets that belong to a specific population of certain body shape and size; therefore, to generate designs for characters of any shape and size, we need to explicitly incorporate the awareness of body shapes and contacts within the generative models. 

On the other hand, while some previous methods \cite{blinn2021learning,mezghanni2022physical} have been optimizing body contact or functionality for objects, their methods are usually limited to common objects. The optimization process does not consider semantics-related design specifications, such as text or image prompts. Additionally, optimizing functionality for creative objects, especially for individual human bodies and preferred contacts, is significantly more challenging and not well-addressed. Meanwhile, there are works that address both body- and semantics-aware objectives, however, they are limited to specific object categories, such as garments \cite{sarafianos2024garment3dgen,wang2018learning}. 

In this work, we propose a tool to generate customized 3D designs that are both body-aware and semantically-aware. The tools can be applied to a wide range of everyday object categories, without relying on object datasets. We build a flexible system that jointly optimizes for multiple objectives. We define semantically-aware design as the process of designing according to a text or visual concept. Personalized, body-aware design is generating a 3D shape that is well-fitted to an individual body or a specific contact map. 

As shown in \Figref{fig:teaser}, we showcase a gallery of our generated designs for various digital avatars and object categories. Our qualitative results show that \model is effective in generating designs that are simultaneously body-aware and semantics-aware. Additionally, we show that compared to baselines, our joint optimization approach achieves the best results in terms of both objective and subjective metrics.

\begin{figure}[t]
    \centering
    \includegraphics[width=0.98\linewidth]{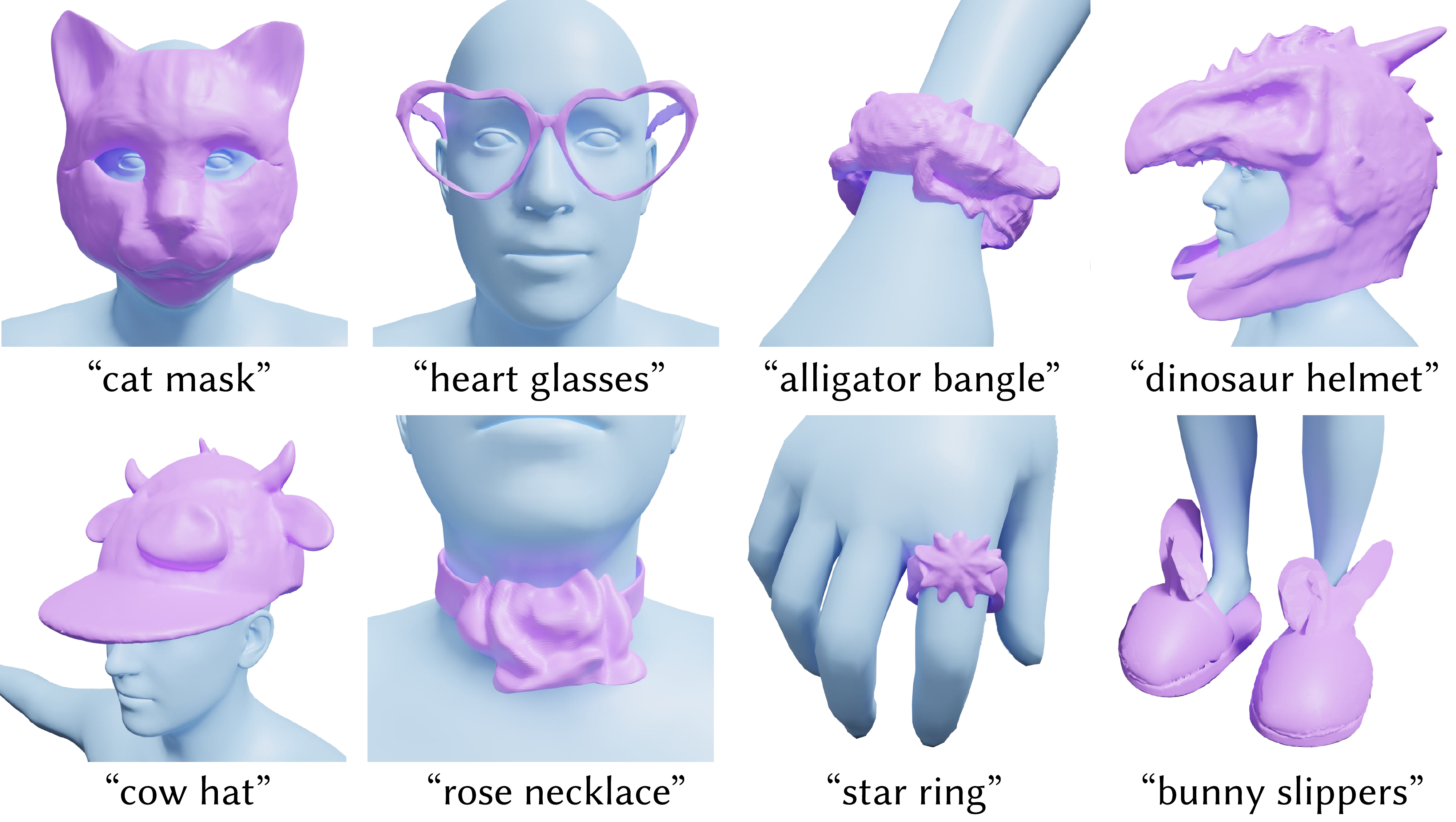}
    \caption{Our method generates a variety of semantics and body-aware objects from input text prompts.}
    \label{fig:01-intro_02-gallery}
\end{figure}

\begin{figure}[t]
    \centering
    \includegraphics[width=0.98\linewidth]{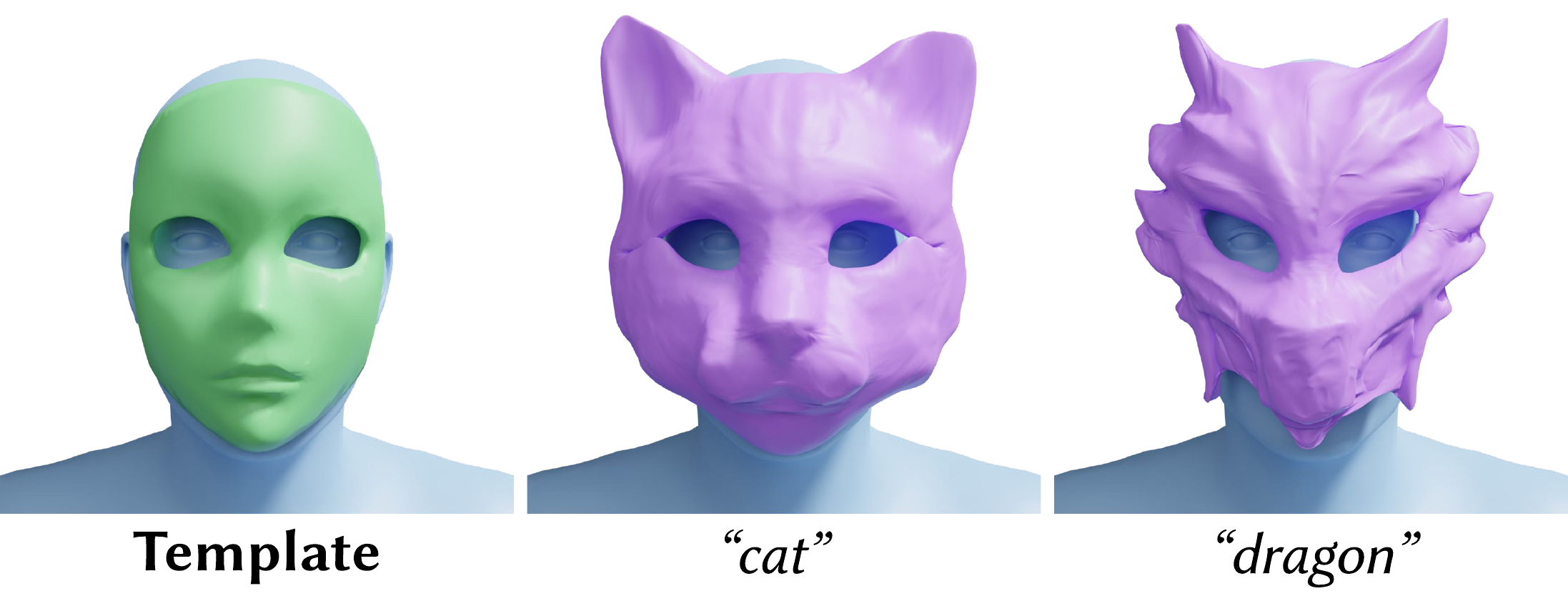}
    \caption{Our method can deform the same template mesh into different text-specified geometries that are body-fitting.}
    \label{fig:04-exp_many-targets}
\end{figure}

\section{Related Work}
\label{sec:related}


\paragraph{Text or image-conditioned 3D synthesis.}
Recent works propose to tackle text-conditioned 3D generation either via text-to-3D~\cite{chen2023fantasia3d,lin2023magic3d,tsalicoglou2023textmesh,zhu2023hifa}, or image-to-3D~\cite{liu2023zero,qian2023magic123,liu2023syncdreamer} where the input image is generated by a text-to-image model such as Stable Diffusion~\cite{rombach2022high} or DALL-E~\cite{dalle3}.
Another line of work directly trains 3D diffusion models for various 3D representations, including point clouds~\cite{nichol2022point}, meshes~\cite{gao2022get3d,liu2023meshdiffusion}, or neural fields~\cite{jun2023shap}.
Finally, other works achieve text or image-conditioned mesh generation by deforming a template mesh through text or image guidance~\cite{michel2022text2mesh,gao2023textdeformer,sarafianos2024garment3dgen}. 

Compared to text-to-image models, text-to-3D is significantly more challenging, partially due to the lack of large-scale training datasets. However, text-to-3D models can leverage pre-trained 2D models, such as CLIP, to synthesize better objects.  Guided by a text prompt (embedded using CLIP), Dreamfields~\cite{dreamfields} synthesize 3D objects leveraging volume rendering.  DreamFusion~\cite{poole2022dreamfusion} and \cite{sjc} distill 2D diffusion models as a differentiable image-based loss. Surface-based differentiable rendering can be used to pass views of explicit 3D objects to CLIP, such as Text2Mesh \cite{michel2022text2mesh} in which they stylize the template mesh while preserving the initial content. CLIP-Mesh~\cite{clipmesh} generates new 3D objects by deforming a sphere at the vertex level, guided by the input text prompt. Magic3D~\cite{lin2023magic3d} first optimizes a radiance field, extracts the mesh from the radiance field, and optimizes the mesh via differentiable surface rendering and score distillation. TextDeformer~\cite{gao2023textdeformer} leverages differentiable rendering and CLIP, but focuses on the problem of deforming explicit geometry rather than generating it from scratch.

\paragraph{Body-aware 3D synthesis.}
The design of 3D objects for human-object interaction is an important research topic. For everyday objects, it is important to consider human bodies, poses, and movements when generating 3D designs for humans \cite{saul2010sketchchair,chen2016reprise}. To optimize for human interaction, various objective functions and evaluation metrics are defined \cite{wu2020chair}. Several previous works have explored this direction, e.g., in 3D room layout generation \cite{sun2024haisor}, scene synthesis \cite{ye2022scene,yi2023mime,vuong2023language,sun2024haisor}, as well as chairs and other body-supporting surfaces design \cite{leimer2018sit,leimer2020pose,zhao2021ergoboss,zheng2015ergonomics,blinn2021learning}. A notable but challenging research direction is garment deformation \cite{kardash2022design,sarafianos2024garment3dgen,wang2018learning,li2023diffavatar}.
To deform 3D objects, one can directly optimize on the 3D space~\cite{sorkine2004laplacian,liu2018paparazzi,jung2024geometry}, using triplanes~\cite{Fruehstueck2023VIVE3D} and text-to-mesh methods~\cite{chen2019text2shape, mohammad2022clip,michel2022text2mesh}. Foundation models can provide guidance or supervision signals for text and image-based stylization~\cite{decatur20233d} and manipulation~\cite{gao2023textdeformer} of 3D objects with various deformation methods~\cite{sumner2004deformation,jacobson2011bounded,baran2009semantic,wang2015linear,zhang2008deformation,gao2018automatic,groueix2019unsupervised,yifan2020neural}. Related effort \cite{richardson2023texture,chen2023text2tex,zeng2023paint3d,yeh2024texturedreamer} applied text-to-image generation models to create textures based on the mesh and given text or image.

\section{Method}
\label{sec:method}

\begin{figure*}[t]
    \centering
    \includegraphics[width=\linewidth]{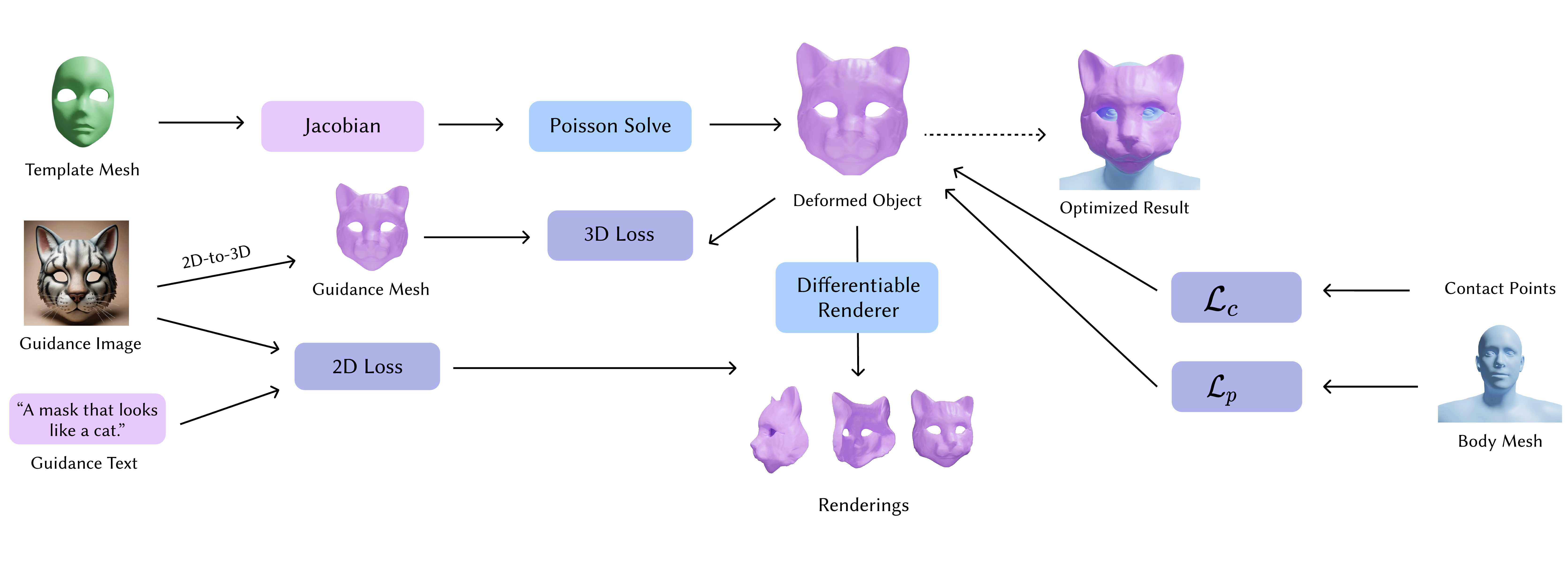}
    \caption{Method overview. We synthesize body-aware 3D objects from a base mesh given an input body geometry and either text or image as guidance. We propose to use a mesh deformation procedure that optimizes for both semantic alignment as well as contact and penetration losses.}
    \label{fig:03-methods_overview}
\end{figure*}

Our goal is to design rigid objects that satisfy diverse contact constraints for different body shapes and semantics. Figure~\ref{fig:03-methods_overview} shows an overview of our method. 
It takes in multiple inputs, including a text prompt or image (e.g., generated by text-to-image models or existing images), a template object mesh, a body mesh, and a set of desired contact points.
We represent the geometry of the input object using a mesh $\mathcal{M}$ with $n$ vertices $\mathcal{V} \in \mathbb{R}^{n\times 3}$ and $m$ faces $\mathcal{F}\in\{1,\ldots,n\}^{m\times 3}$.
We aim to optimize a displacement map $\Phi:\mathbb{R}^3\to\mathbb{R}^3$ across the vertices.

\paragraph{Shape optimization through Jacobians.}

The design parameterization plays a significant role in the difficult design optimization problem. 
Naive optimization of the mesh deformation through vertex displacement can result in significant artifacts and is prone to convergence to local minima~\cite{gao2023textdeformer}. Inspired by Neural Jacobian Fields~\cite{aigerman2022neural}, we indirectly optimize the deformation map by optimizing a set of per-triangle Jacobian matrices $J_i\in\mathbb{R}^{3\times 3}$ for every face $f_i\in\mathcal{F}$. 
The deformation map $\Phi^\ast$ is computed as the mapping with Jacobian matrices that are closest to $\{J_i\}$, solved via the following Poisson optimization problem:
\begin{equation}
    \Phi^\ast = \min_{\Phi} \sum_{f_i\in\mathcal{F}}|f_i|\lVert \nabla_i(\Phi) - J_i\rVert_2^2,
    \label{eq:poisson_equation}
\end{equation}
where $\nabla_i(\Phi)$ denotes the Jacobian of $\Phi$ at triangle $f_i$ and $|f_i|$ is the area of that triangle.

\subsection{Semantics-Aware Optimization}

The user has the option to specify the semantic goals with a text prompt or an input image. Depending on the input modality, our system uses different losses to guide the optimization. We describe the losses for each modality below.

\paragraph{Input text guidance.}
For text guidance, the goal during the optimization process is to ensure that the resulting object aligns with the text prompt that specifies the desired design outcome. The pre-trained CLIP~\cite{radford2021learning} provides a joint text-image feature space, which can be used for this alignment objective. We pass the current deformed mesh $\Phi^\ast(\mathcal{M})$ to a differentiable renderer $\mathcal{R}$~\cite{nvdiffrast} to generate $K$ images from different views:
\begin{equation}
    I_k = \mathcal{R}(\Phi^\ast(\mathcal{M})), \;\;\; k=1,\ldots, K.
\end{equation}
The images are passed to CLIP to obtain the embeddings of the renders $\texttt{CLIP}\big( I_k \big) \in \mathbb{R}^{512}$. We pass the text prompt $\mathcal{P}$ to CLIP to get the language embedding with the same dimension, $\texttt{CLIP}(\mathcal{P}) \in \mathbb{R}^{512}$. Then, we define the text alignment objective to be the negative cosine similarity between the embeddings:
\begin{equation}
    \mathcal{L}_s(\mathcal{M}) = \frac{1}{K}\sum_{k=1}^K-\mathrm{sim}\left(\texttt{CLIP} \big( I_k \big), \texttt{CLIP} \big( \mathcal{P} \big)\right).
    \label{eq:identity}
\end{equation}
Since CLIP operates on 2D images, multi-view consistency is a challenge. Averaging gradients across different views of the object often results in inconsistent artifacts such as incorrect geometry. We adopt the regularization term developed in \cite{gao2023textdeformer}, which tackles this problem by utilizing the patch-level deep features of CLIP's vision transformer (ViT). The intuition is that we can split the image into small patches, which are then projected into a higher-dimensional space. For each vertex and each render, we compute the pixel value in that render that contains the vertex. Then, by by associating the pixel value with the nearest corresponding patch center, we obtain a feature vector for that vertex in that render. In this way, we can encourage vertices to have similar deep features across renders from different viewpoints.

\paragraph{Input image guidance.}
If the user provides an input image $\bar{I}$, the goal of the optimization process is to optimize the shape of the object such that it matches the design in the input image.
Inspired by Sarafianos et al.~\cite{sarafianos2024garment3dgen}, we use an image-to-3D model~\cite{tripoai} to lift the image to a 3D guidance mesh denoted as $\overline{\mathcal{M}}$. Similar to text guidance, we render the guidance mesh from the $K$ different views:
\begin{equation}
    \bar{I}_k = \mathcal{R}(\overline{\mathcal{M}}), \;\;\; k=1,\ldots, K,
\end{equation}
and compute the cosine similarity of the CLIP embeddings of the guidance mesh renders and the current deformed mesh, averaged across the views:
\begin{equation}
    \mathcal{L}_s(\Phi^\ast(\mathcal{M}), \overline{\mathcal{M}}) =  \frac{1}{K}\sum_{k=1}^K -\text{sim} \left( \texttt{CLIP} \big( I_k \big), \texttt{CLIP} \big(\bar{I}_k \big) \right).
\end{equation}
This loss acts as a soft constraint between the embeddings of the deformed mesh $\Phi^\ast(\mathcal{M})$ and those of the pseudo-ground truth $\overline{\mathcal{M}}$. For stronger 3D supervision, we use a two-sided Chamfer Distance (CD) loss to measure the distance between two sets of points, \(p \in S\) and \(\bar{p} \in \bar{S}\), sampled from $\Phi^\ast(\mathcal{M})$ and $\overline{\mathcal{M}}$, respectively, in each optimization step:
\begin{equation}
    \mathcal{L}_\text{CD} = \frac{1}{|S|}\sum_{p \in S} \min_{{\bar{p}} \in \bar{S}} \lVert p - \bar{p}\rVert_2^2 + \frac{1}{|\bar{S}|}\sum_{\bar{p} \in \bar{S}} \min_{{p} \in S} \lVert \bar{p} - p\rVert_2^2.
\end{equation}
For 2D supervision, we use an L1 loss to ensure that the deformed mesh does not deviate too much from the image guidance along each step of the optimization:  
\begin{equation}
    \mathcal{L}_\text{2D} = \frac{1}{K}\sum_{k=1}^K \lvert I_k - \bar{I}_k \rvert.
\end{equation} 

\subsection{Body-Aware Optimization}

\begin{figure}[t]
    \centering
    \includegraphics[width=0.98\linewidth]{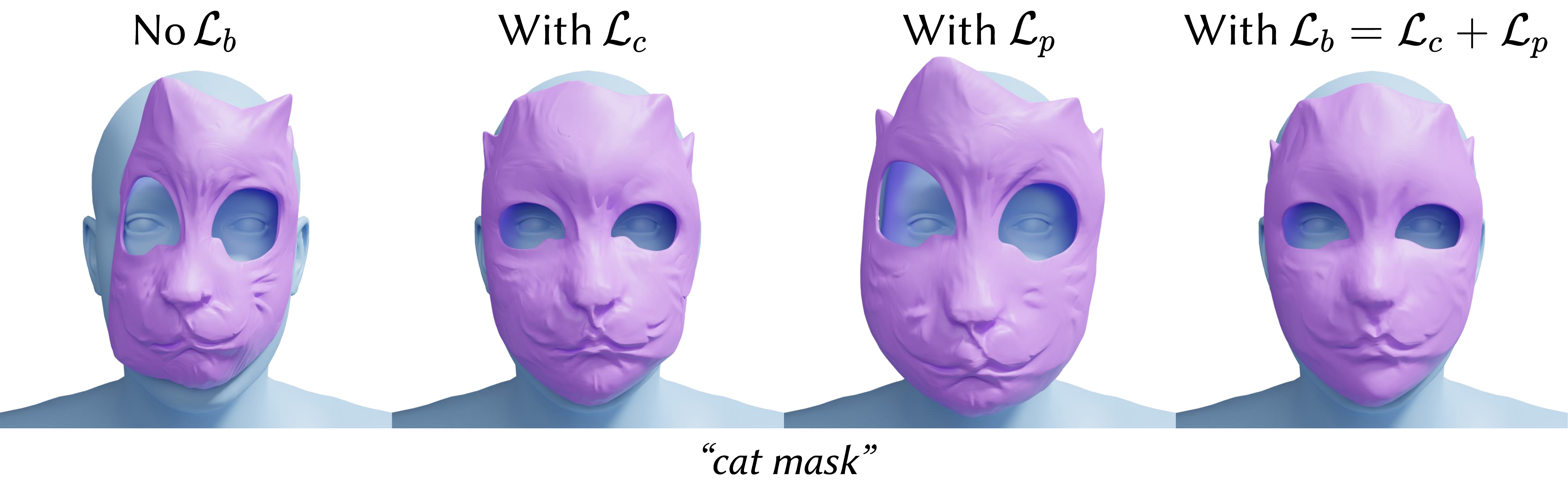}
    \caption{We show the effect of contact vs. penetration losses on text guided deformation for ``cat mask''.}
    \label{fig:03-methods_bp-ablation-text}
\end{figure}

A key component of our optimization procedure is to produce objects that will satisfy contact constraints for different body shapes. Inspired by \cite{ye2022scene}, given contact vertices $\mathcal{V}_c$, the contact loss $\mathcal{L}_c$ is defined as
\begin{equation}
    \mathcal{L}_c(\mathcal{V}, \mathcal{V}_c) = \lambda_c\frac{1}{|\mathcal{V}_c|} \sum_{v_c \in \mathcal{V}_c} \min_{v \in \mathcal{V}} || v_c - v||^2_2,
    \label{eq:contact_loss}
\end{equation}
where $\lambda_{contact}$ is a tunable weight. This encourages the object to be in contact with the body vertices specified by the input contact vertices. To reduce penetration between the object and the body mesh $\mathcal{M}_b$, we include an additional loss $\mathcal{L}_p$:
\begin{equation}
    \mathcal{L}_p(\mathcal{M}, \mathcal{M}_b) = \sum_{d_i < D} {d_i}^2,
    \label{eq:pen_loss}
\end{equation}
where $d_i$ are signed distances between the object and the body mesh, and $D$ is the penetration distance threshold. 
In total, the body-aware optimization loss is defined as:
\begin{equation}
    \mathcal{L}_b(\mathcal{V}, \mathcal{V}_c, \mathcal{M}, \mathcal{M}_b) = \lambda_c \mathcal{L}_c(\mathcal{V}, \mathcal{V}_c) + \lambda_p \mathcal{L}_p(\mathcal{M}, \mathcal{M}_b).
\end{equation}
In \Figref{fig:03-methods_bp-ablation-text}, we show the effect of the contact loss $\mathcal{L}_c$ and the penetration loss $\mathcal{L}_p$ during the deformation procedure for the text prompt ``a mask that looks like a cat''.
While semantic optimization severely penetrates the face, integrating contact and penetration losses improve the fit and reduce the penetration, respectively.

\subsection{Optimization Problem Statement}
\begin{figure}[t]
    \centering
    \includegraphics[width=0.99\linewidth]{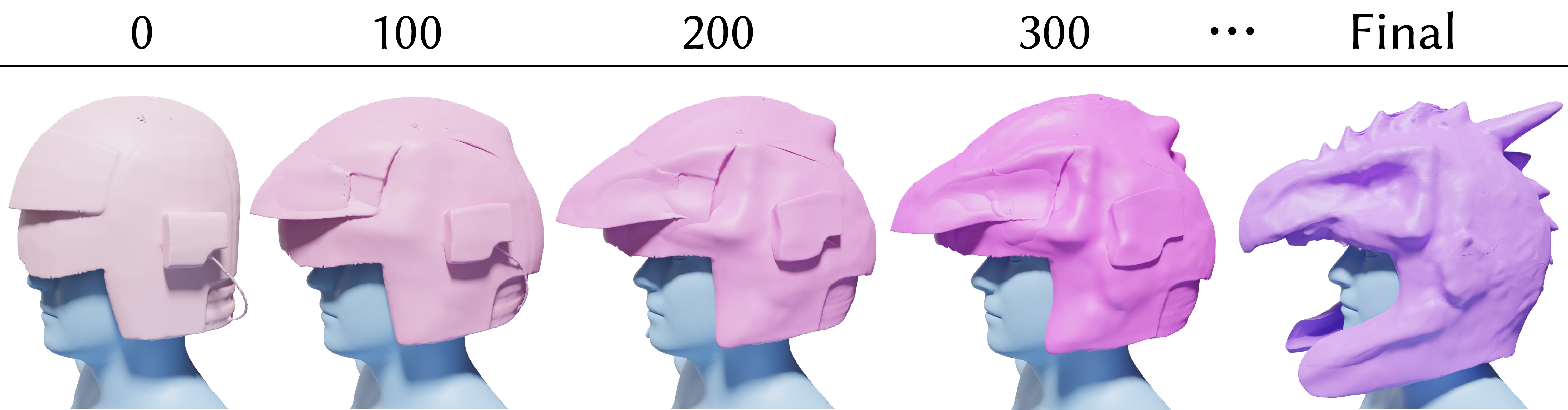}
    \caption{Evolution of design throughout optimization iterations.}
    \label{fig:03-methods_evolution}
\end{figure}

In summary, the optimization objective is to optimize the Jacobian matrices ${J_i}$ according to the weighted sum of the semantics-aware and body-aware losses:
\begin{align}
    \mathcal{L}(\mathcal{M}) &= \lambda_s\mathcal{L}_s(\Phi^\ast(\mathcal{M}), \overline{\mathcal{M}}) \\
    &+ \lambda_b\mathcal{L}_b(\mathcal{V}, \mathcal{V}_c, \mathcal{M}, \mathcal{M}_b) \\
    &+ \alpha\sum_{i=1}^{|\mathcal{F}|} \lVert J_i - \bm{I}\rVert_2.
\end{align}
The last term regularizes the predicted Jacobians, where $\bm{I}$ denotes the identity matrix, and $\alpha$ controls the strength of the deformations.
We show the evolution of the mesh deformation process across optimization iterations in \Figref{fig:03-methods_evolution}.
During the optimization process, the mesh becomes more semantically aligned with the input guidance, while fitting the body well.

\section{Experiments and Results}

In our experiments, we seek to answer the following questions:
\begin{itemize}[leftmargin=12pt,noitemsep]
    \item Can \model be used to generate object designs across different semantic targets and body shapes?
    \item What is the effect of the choice of guidance (text vs. image)?
    \item How does the body loss affect the design and fit of the object on the body?
    \item Is joint optimization better than two-stage optimization?
    \item How does our method compare with baseline methods in terms of semantic alignment and body fit?
\end{itemize}

\subsection{Generality of \model}
\paragraph{Different object categories and design specifications.}
As shown in \Figref{fig:01-intro_02-gallery}, \model is a general method that can generate semantically and body-aware everyday objects. Here we cover a variety of objects that need to be attached to different parts of the body: head (mask, glasses, helmet, visor), neck (necklace), wrist (bangle), finger (ring), and foot (slippers). 

Next, we assess whether the same base mesh can deform into multiple target prompts in \Figref{fig:04-exp_many-targets}. We find that indeed the same base mesh can be used for different text prompts within the same object category, given that the topology within an object category are often shared across designs.

\begin{figure}[t]
    \centering
    \includegraphics[width=0.98\linewidth]{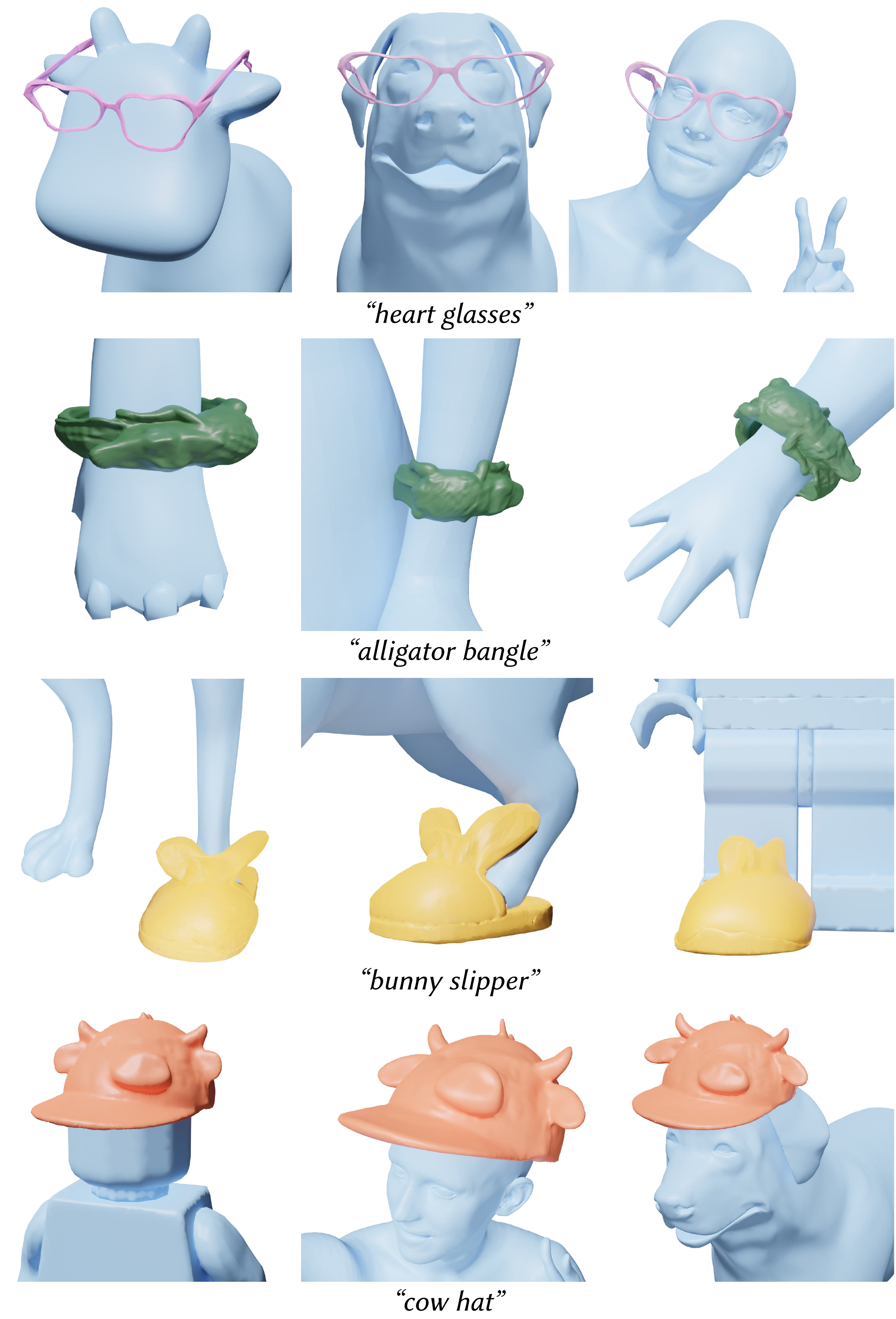}
    \caption{Our method can customize the same object design for different character body shapes.}
    \label{fig:04-exp_characters}
\end{figure}

\paragraph{Different body shapes.}
We evaluate our system on different body shapes, ranging from human adults and children to virtual characters such as dinosaurs and cartoon-looking cows.
\Figref{fig:04-exp_characters} shows the same text prompt for different character body shapes. We observe that different bodies affect the creativity of the optimization due to the amount of free space the object has to deform on the character without penetrating the body, but \model is able to optimize for individual body shapes. Given the prompt ``bunny slipper'', we show the optimization results for dog, dinosaur, and LEGO characters in \Figref{fig:04-exp_characters} (third row). We observe that the slipper's ears vary across the body shapes with noticeable differences in length and orientation. For example, the slipper ears are much more pronounced on the dog than on the LEGO character. This is due to the variance of body shapes -- the dog's thin leg provides more room for the slipper to grow. Even though the rigid leg on the LEGO character prevents the bunny ears from growing longer, it still manages to be prompt-aligned.

\subsection{Justification of Design Choices}

\begin{figure}[t]
    \centering
    \includegraphics[width=0.98\linewidth]{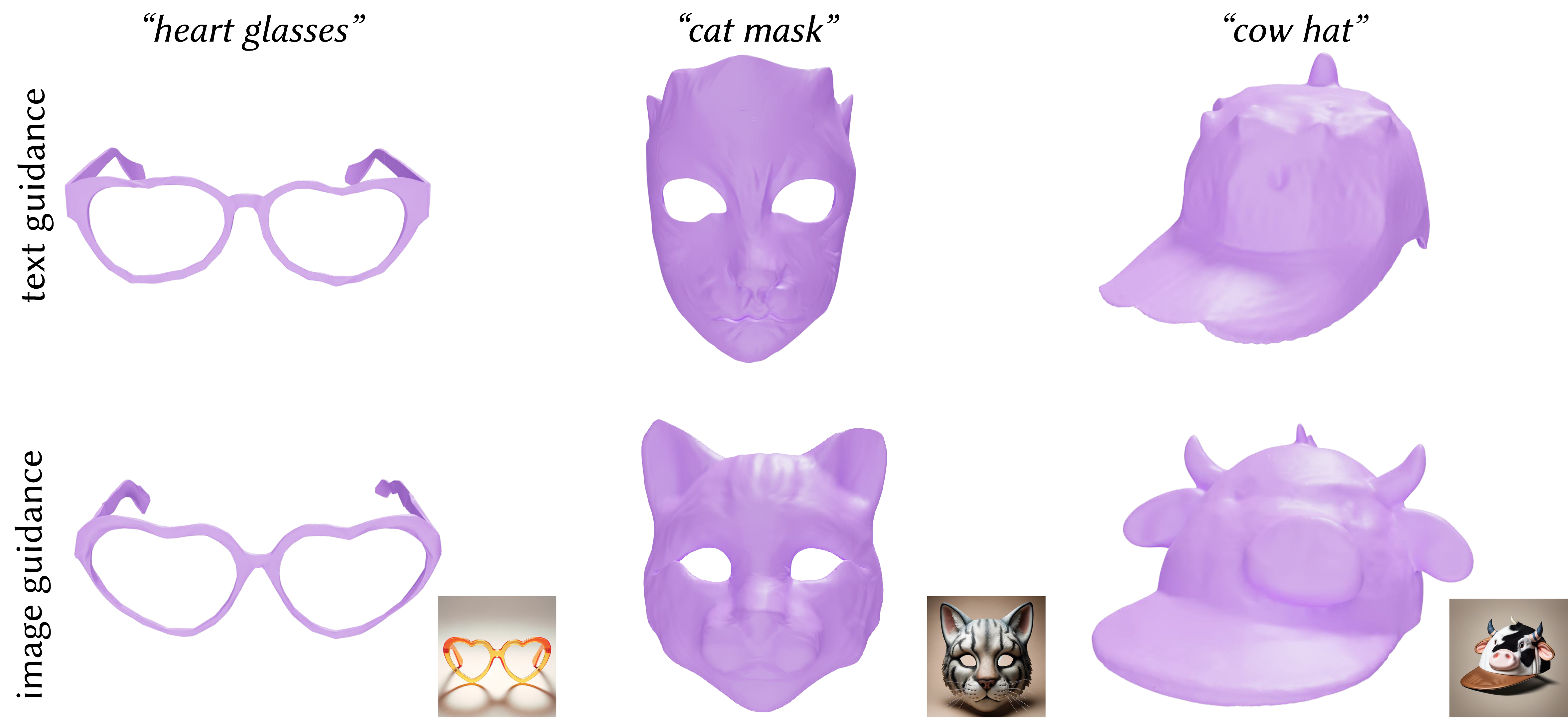}
    \caption{We evaluate the effect of text vs. image guidance. Image guidance produces stronger control, generating objects that are more prompt-aligned. We show the reference image in the bottom right corner of example of the image guidance row.}
    \label{fig:04-exp_guidance-ablation}
\end{figure}

\paragraph{Effect of text vs. image guidance.}
We evaluate the effect of text vs. image guidance in \Figref{fig:04-exp_guidance-ablation}.
While text guidance occasionally produces the desired semantics (e.g., ``heart glasses'' and ``cat mask''), it exhibits limited deformation on most examples (e.g., ``star ring''). In contrast, image guidance provides a stronger signal for deformation.

\begin{figure}[t]
    \centering
    \includegraphics[width=0.98\linewidth]{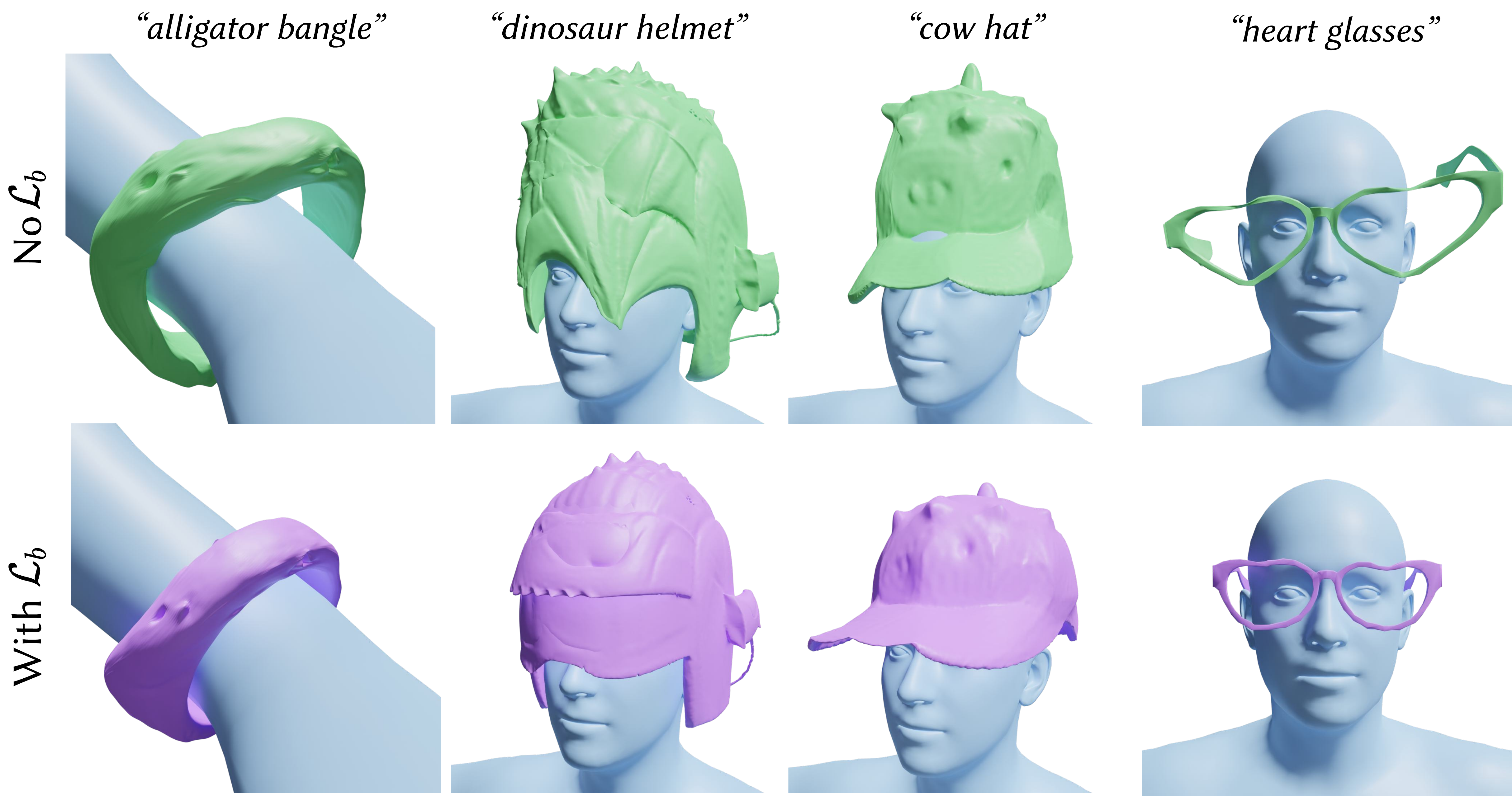}
    \caption{With contact and penetration losses, the text-guided deformations are more body fitting.}
    \label{fig:04-exp_body-ablation-text}
\end{figure}

\paragraph{Effect of body loss.}
In \Figref{fig:04-exp_body-ablation-text}, we analyze how body losses affect the fit of the object on the body. Although the initial base mesh starts off fitting well, it either penetrates or loses contact with the body when optimizing for semantic alignment. Incorporating the body loss alleviates this issue.

\begin{figure}[t]
    \centering
    \includegraphics[width=0.98\linewidth]{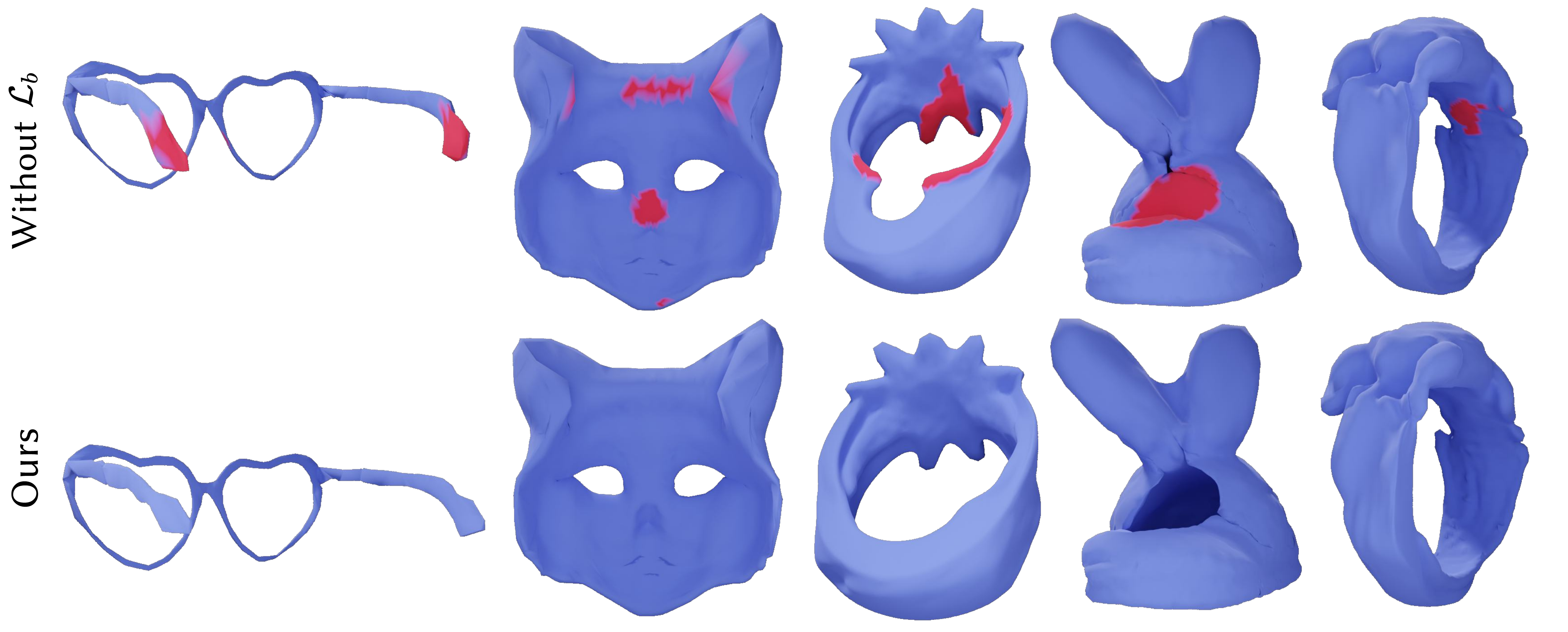}
    \caption{We visualize penetration maps on objects optimized with (second row) and without body losses (first row). In the penetration maps, the blue regions indicate a positive distance between the mesh and the characters, signifying no penetration. Red regions indicate a negative distance, denoting penetration between the mesh and the interacting character. Without the incorporation of body losses, the generated objects exhibit significant penetration with the character.}
    \label{fig:04-exp_body-ablation-img}
\end{figure}

In \Figref{fig:04-exp_body-ablation-img}, we compare objects that are optimized using image guidance, with and without the body loss. We find that including the body loss in the image guidance optimization helps minimize penetrations between the object and the human.

\paragraph{Comparison with two-stage optimization.}

\begin{figure}[t]
    \centering
    \includegraphics[width=0.98\linewidth]{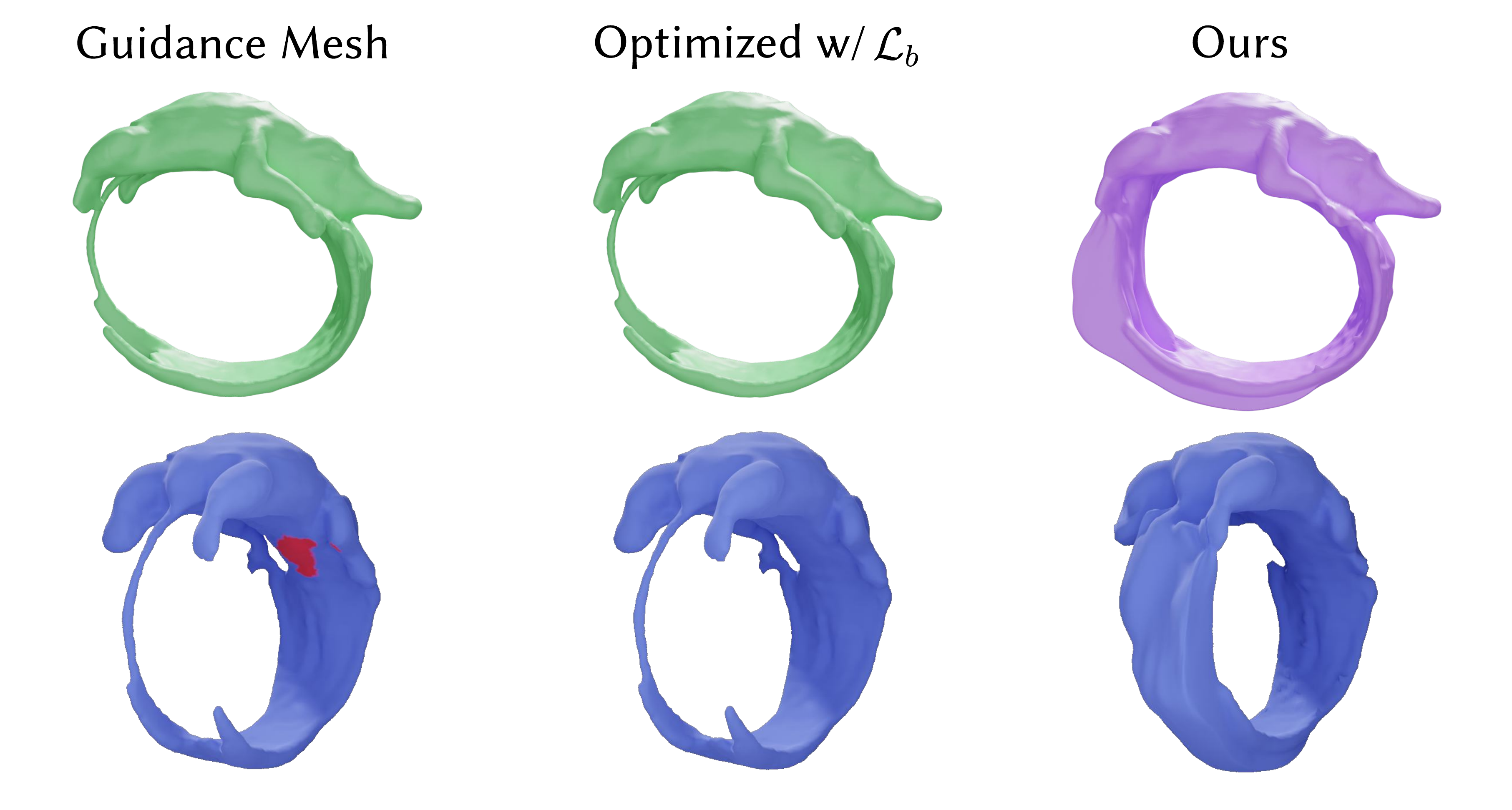}
    \caption{We show the object mesh (first row) and the penetration map from a different viewpoint (second row). Even though we can apply a body refinement optimization on the guidance mesh to reduce the penetrations, it cannot the fix the thin structure on the object. Jointing optimizing for both body and semantics together results in a more well-formed mesh while also minimizing penetrations.}
    \label{fig:04-exp_fix-thin-structures}
\end{figure}

In \Figref{fig:04-exp_fix-thin-structures}, we analyze alternatives for image-guided 3D generation: (i) the guidance mesh (generated via text-to-image and image-to-3D models), (ii) the guidance mesh with a second body-aware refinement stage.
Although making the guidance mesh body-aware reduces penetrations, it cannot make the thin parts of the guidance mesh wider. 
In contrast, jointly optimizing for both body and semantics from a template mesh results in a fitting bangle that is aligned with the prompt.

\begin{figure}[t]
    \centering
    \includegraphics[width=0.98\linewidth]{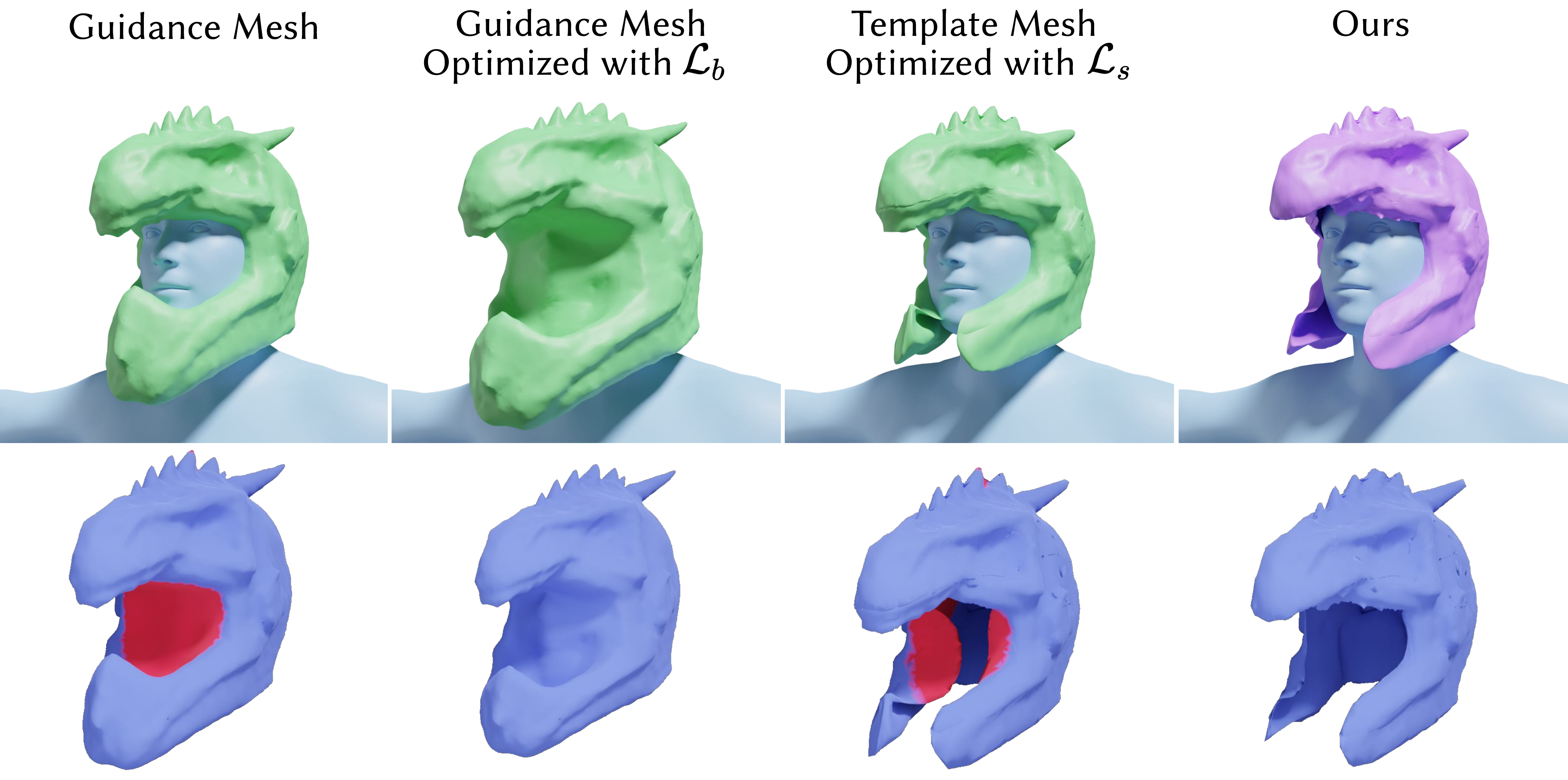}
    \caption{We show the object mesh (first row) and the penetration map (second row). When the guidance mesh lacks the correct topology, such as missing a hole for the head in the helmet, body refinement cannot fix the issue and ends up enclosing the entire head. Starting from the template mesh and optimizing for semantics alone is also not sufficient; while the topology is correct, the optimizing may introduce penetrations.}
    \label{fig:04-exp_fix-topology}
\end{figure}

In \Figref{fig:04-exp_fix-topology}, we analyze another example with a ``dinosaur helmet''. Because the topology of the guidance mesh is incorrect (missing holes for the head), applying a body-aware refinement step optimizes penetrations by enclosing the entire head. Starting from the base mesh and optimizing for semantics alone is also not sufficient: despite having the correct topology, the optimization may introduce penetrations. Our method is able to properly generate a helmet that is prompt-aligned while maintaining a hole for the human's head.

\subsection{Comparison with Baselines}
To provide a quantitative evaluation of \model against baselines, we compute prompt alignment, contact distance, and penetration distance on all the different method variants. As shown in \Tabref{tab:baselines}, we find that incorporating body loss significantly reduces penetration, regardless of the base mesh used. Furthermore, the Chamfer distance of the contact points are also well-maintained by using body loss.

\begin{table}[t]
\footnotesize
\caption{We report quantitative metrics on the prompt alignment, penetration, and chamfer distance of the contact points to the object vertices.}
\begin{tabular}{llllccc}
\hline
Base Mesh & Input Modality & $\mathcal{L}_b$ & CLIP $\uparrow$ & $D_p$ $\downarrow$ & $D_c$  $\downarrow$ \\ \hline
Template & N/A & N/A & 0.25 & 6.7e-2 & 1.9e-10 \\
Guidance & N/A & N/A & 0.28 & 65.6 & 2.1e-10 \\
Guidance & N/A & Y & 0.27 & 8.5e-4 & 5.1e-3 \\
Template & Image & N & 0.27 & 3.7 & 7.6e-3 \\
Template & Image (Ours) & Y & 0.27 & 7.0e-4 & 4.6e-3 \\ \hline
\end{tabular}
\label{tab:baselines}
\end{table}

We also conduct a user study (N=9) asking participants to rate (on a Likert scale of 1-10) the prompt alignment (``how aligned/similar are each of objects to the original prompt?''), aesthetics (``how aesthetic are the following objects?''), and the perceived comfort of the object on the human body (``how comfortable do the following objects look on the human?''). Our method performs the best on prompt alignment and aesthetics, and achieves comparable performance on comfort when compared with the template mesh. We see that the template mesh achieves the lowest score on prompt alignment, which is expected, because the template mesh is only representative of the object category, and not adapted to the creative prompt. We see that TextDeformer itself has the lowest score in comfort which is expected, as the objects are optimized without considering body fit. Although Body-Aware TextDeformer achieved a higher score in comfort compared to TextDeformer, its alignment score decreased. This indicates that the deformation prioritized body-awareness in a way that conflicted with the object's prompt alignment. In contrast, our method remains the most prompt aligned and aesthetic across all methods while also maintaining comfort, showing our method successfully prioritizes both body-awareness and semantic-awareness.
\begin{table}[]
\small
\centering
\caption{We report results on the user study. We ask participants to rate the prompt alignment, aesthetics, and the perceived comfort of generated objects on the body.}
\begin{tabular}{@{}lccc@{}}
\toprule
 & Align. $\uparrow$ & Aesth. $\uparrow$ & Comf. $\uparrow$ \\ \midrule
Template mesh & 1.47 & 5.25 & \textbf{6.74} \\
TextDeformer~\cite{gao2023textdeformer} & 3.19 & 2.94 & 2.88 \\
Body-Aware TextDeformer & 3.04 & 3.96 & 5.17 \\
Ours & \textbf{7.78} & \textbf{6.40} & 5.75 \\ \bottomrule
\end{tabular}
\label{tab:user}
\end{table}

\subsection{Applications of \model}
\begin{figure}[t]
    \centering
    \includegraphics[width=0.98\linewidth]{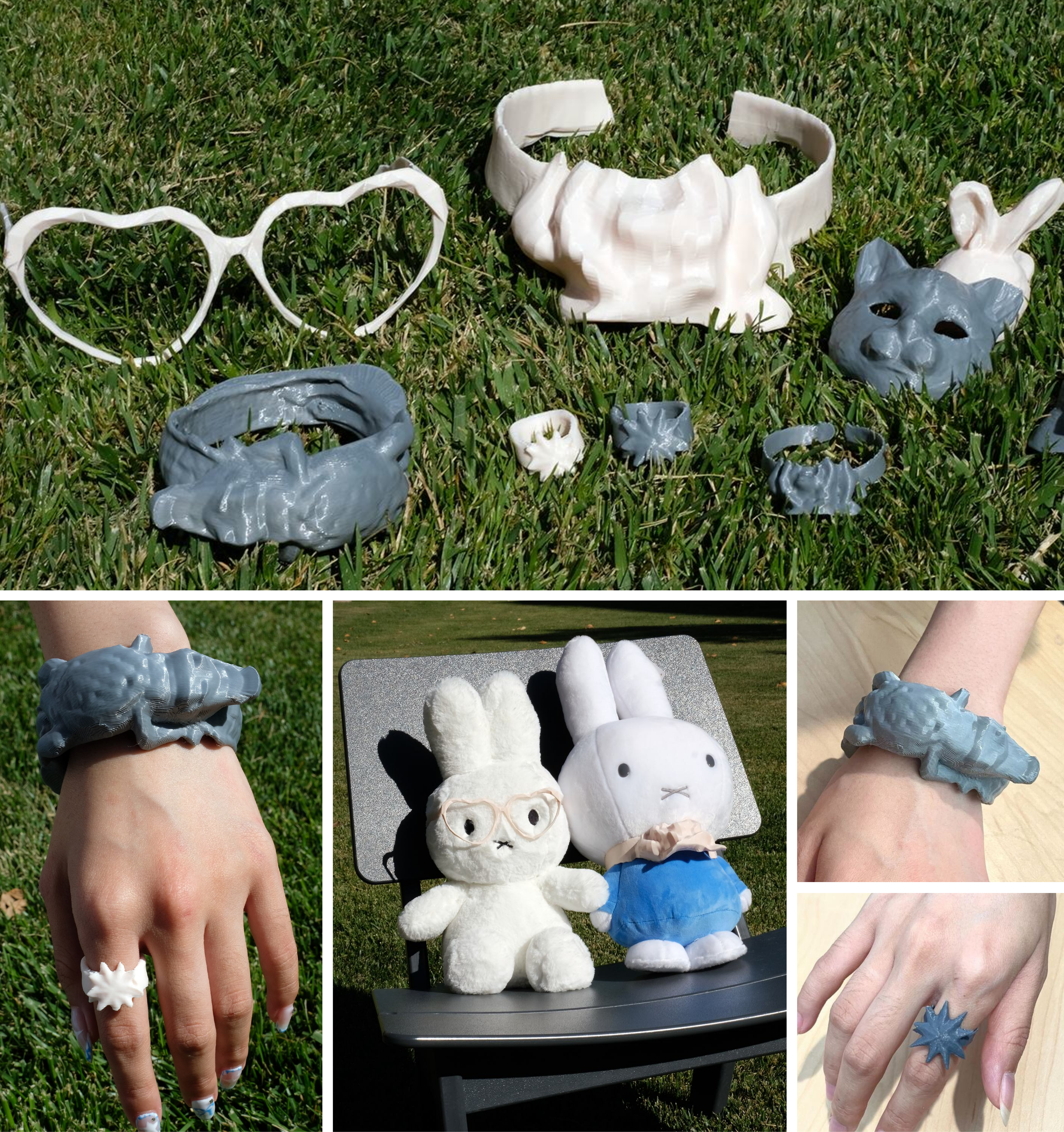}
    \caption{We fabricated the objects in the real world with 3D printing. The objects can be worn as accessories on real people and characters.}
    \label{fig:04-exp_04-fabrication}
\end{figure}

\paragraph{Fabricated designs.}
The designs generated by our system are fabricable in the real world. In \Figref{fig:04-exp_04-fabrication} (left), we show the  objects 3D-printed using \model-generated meshes without manual modifications. In \Figref{fig:04-exp_04-fabrication} (right) we also show how objects fit on a real human body and characters.

\begin{figure}[t]
    \centering
    \includegraphics[width=0.98\linewidth]{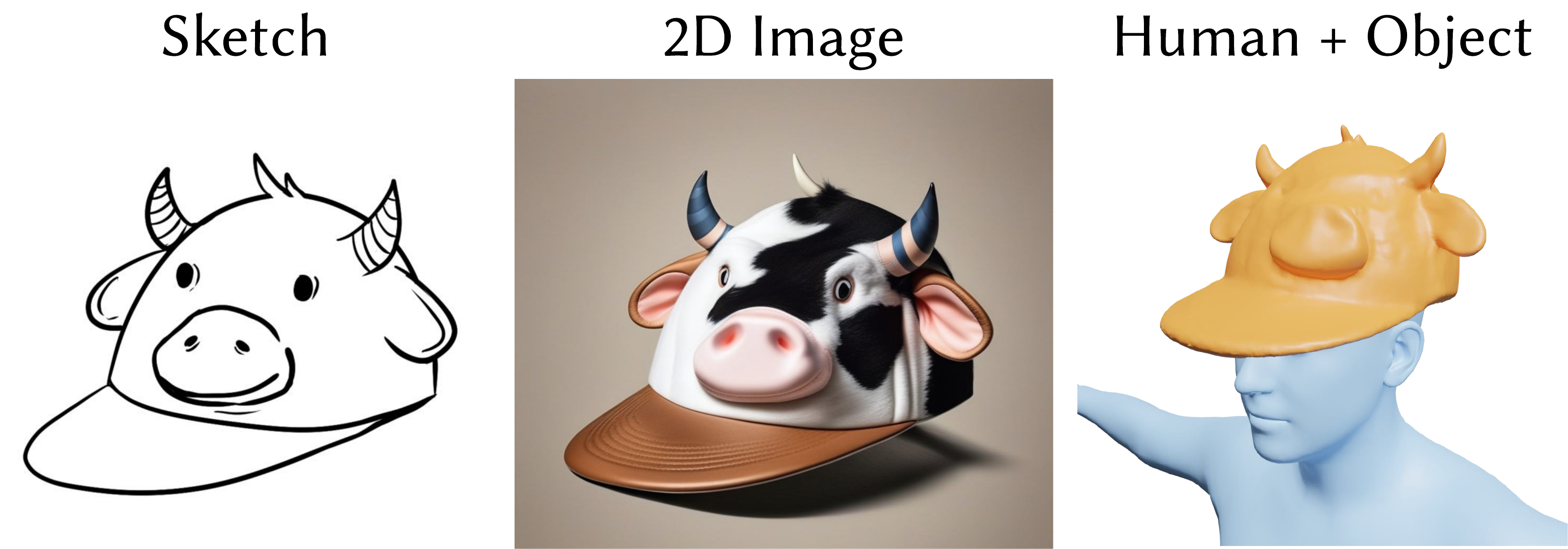}
    \caption{We show an application of lifting a sketch into a body-fitting 3D object design.}
    \label{fig:04-exp_sketch}
\end{figure}

\paragraph{Sketch-guided design.}
We show a sketch application with \model.
We ask a user to draw a sketch of an object, and we use ControlNet~\cite{zhang2023adding} to convert the sketch into a 2D image. The image is used as an input to our image-guided mesh deformation method. The results are shown in \Figref{fig:04-exp_sketch}.

\section{Conclusion}

In this work, we present \model, a 3D object design framework that integrates body and semantic awareness into the generative process. Our method synthesizes body-aware 3D objects from a base mesh using input body geometry and guidance from text or images. The joint optimization for semantic alignment and body-aware losses ensures that the generated objects are both creatively customized and functionally practical. Our evaluations demonstrate the efficacy of ShapeCraft in producing virtual and real-world objects that fit a wide range of body shapes without the need for manual intervention. ShapeCraft not only streamlines the design process but also enables the fabrication of personalized, body-aware objects, thereby enhancing customization and usability in everyday object design.

\paragraph{Acknowledgments.}
This work is in part supported by the Stanford Institute for Human-Centered AI (HAI), the Stanford Center for Integrated Facility Engineering (CIFE), Nissan, and NSF RI \#2338203. MG is supported by NSF GRFP. Ruohan Zhang is partially supported by Wu Tsai Human Performance Alliance Fellowship.

{\small
\bibliographystyle{ieee_fullname}
\bibliography{main}
}

\end{document}